%% file: main.tex
\documentclass[lnbip]{svmultln}
\usepackage{makecell}
\usepackage{longtable}
\usepackage[nocompress]{cite}
\usepackage{makeidx}  
\usepackage{xcolor}
\usepackage{graphicx}
\usepackage{subcaption}

\begin{document}
\mainmatter              

\title{Just Tell Me: Prompt Engineering in \\ Business Process Management}
\titlerunning{Just Tell Me: Prompt Engineering in BPM}  
%
\author{Kiran Busch\inst{1} \and Alexander Rochlitzer\inst{1} \and Diana Sola\inst{2,3} \and Henrik Leopold\inst{1}}
\authorrunning{Busch et al.}   
%

%
\institute{Kühne Logistics University, Hamburg, Germany \and SAP Signavio, Walldorf, Germany \and Data and Web Science Group, University of Mannheim, Mannheim, Germany} 
\maketitle              

\vspace{-1em}
\begin{abstract}        

\input{sections/abstract}

\end{abstract}
\vspace{-2.6em}
\section{Introduction}
\input{sections/introduction}

\vspace{-0.8em}
\section{Background}
\label{sec:background}
\input{sections/background}
\vspace{-2em}
\section{Potentials}
\label{sec:potentials}
\input{sections/potentials}

\vspace{-1em}
\section{Challenges}
\label{sec:challenges}
\input{sections/challenges}
\vspace{-1em}
\section{Conclusion} 
\label{sec:conclusion}
\input{sections/conclusion}

\end{document}

%% file: sections/abstract.tex
GPT-3 and several other language models (LMs) can effectively address various natural language processing (NLP) tasks, including machine translation and text summarization. Recently, they have also been successfully employed in the business process management (BPM) domain, e.g., for predictive process monitoring and process extraction from text. This, however, typically requires fine-tuning the employed LM, which, among others, necessitates large amounts of suitable training data. A possible solution to this problem is the use of prompt engineering, which leverages pre-trained LMs without fine-tuning them. Recognizing this, we argue that prompt engineering can help bring the capabilities of LMs to BPM research. We use this position paper to develop a research agenda for the use of prompt engineering for BPM research by identifying the associated potentials and challenges.

%% file: sections/introduction.tex
The recent introduction of ChatGPT has dramatically increased public awareness of the capabilities of transformer-based language models (LMs). However, already for a while, LMs are used to address several common natural language processing (NLP) tasks including search, machine translation, and text summarization. Also in the business process management (BPM) community, LMs have been used for tasks such as process extraction from text~\cite{bellan2022extracting} or activity recommendation~\cite{sola2023pretrained}. 
To accomplish this, pre-trained models are typically fine-tuned, which transforms the pre-trained LM into a task-specific model. The performance of fine-tuning, however, is highly dependent on the amount and quality of downstream data available, which is a common issue in BPM practice~\cite{kappel2021evaluating,bellan2022extracting}.

Recent studies have shown that prompt engineering~\cite{liu2023pre}, which leverages pre-trained LMs without fine-tuning them, can effectively address the issue of limited downstream data and yield promising results in various NLP tasks~\cite{brown2020language,liu2019text,wang2019learning} and in other domains like reasoning~\cite{kojima2022large}. Prompt engineering involves the use of natural language task specifications, known as prompts, which are given to the LM at inference time to provide it with information about the downstream task.
For example, when creating a prompt for the extraction of a topic from the text ``Eliud Kipchoge is a Kenyan long-distance runner'', we could append ``This text is about \rule{0.5cm}{0.4pt}?'' to the text and ask the LM to fill the blank with a topic. Thus, without altering the model itself, prompts provide the context for a downstream task and enable the customization of the outputs and interactions with an LM. 

Research towards prompt engineering for BPM tasks is, however, still in its early stages. To our knowledge, only Bellan et al.~\cite{bellan2022extracting} have conducted research in this field, with the aim of extracting process information from text. While their study demonstrated the potential of using conceptual definitions of business process entities to set the context for the extraction task, the authors acknowledged that providing an LM with the appropriate contextual BPM knowledge via prompts may be a challenging problem that requires further investigation. Against this background, we use this position paper to promote the use of prompt engineering for BPM research. Specifically, we identify potentials and challenges pertaining to the use of prompt engineering for BPM tasks. We believe that prompt engineering has the potential to effectively address a large variety of NLP-related BPM tasks and, hence, reduce the need for highly specialized and use-case specific techniques as well as the need to obtain large training datasets.

%% file: sections/background.tex
Recent advances in the NLP field have led to powerful LMs, which have shown remarkable capabilities across a diverse range of tasks such as text summarization~\cite{liu2019text}, 
machine translation~\cite{wang2019learning}, 
reasoning~\cite{kojima2022large}, and many more.
The success of applying such models to downstream tasks can be attributed to the transformer architecture~\cite{vaswani2017attention} and increased model sizes in combination with the computational capacity of modern computer systems, as well as the models' pre-training on massive volumes of unlabeled text.
Pre-training an LM 
enables it to develop general-purpose abilities that can be transferred to downstream tasks \cite{raffel2020exploring}.

The traditional approach for performing a downstream task with a pre-trained LM is \textit{fine-tuning},
which involves updating the LM's parameters by training it on a large dataset of labeled examples specific to the downstream task~\cite{brown2020language}. 
For example, when fine-tuning an LM for the activity-identification task in the context of transforming a text into a process model \cite{van2018challenges}, such a labeled example could be the (input text, activity)-pair (``He ordered the shoes'', ``order shoes''). 
However, a major limitation in the fine-tuning paradigm is the need for large task-specific datasets and task-specific training~\cite{brown2020language}, which has two drawbacks. First, the performance of fine-tuning is greatly impacted by the number and quality of examples that are specific to the task, while in practice, the scenarios with sparse data are common. Second, fine-tuning modifies all parameters of the LM, thus requiring a copy of the model to be stored for each task.

Therefore, we are currently observing a paradigm shift from fine-tuning to \textit{prompt engineering}~\cite{liu2023pre}, which is driven by the remarkable task-agnostic performance of LMs~\cite{brown2020language,schick2020s}. 
In the prompt-engineering paradigm, natural language task specifications, referred to as prompts, are provided to the LM at inference time to set the context for a downstream task without changing the LM itself. 
This approach of ``freezing'' pre-trained models is particularly attractive, as model sizes continue to increase. 
In certain situations, using pre-trained models in combination with prompt engineering has been demonstrated to be competitive with or even better than state-of-the-art fine-tuned models~\cite{brown2020language,schick2020s,shin2020autoprompt}.

Prompt engineering is typically implemented in a zero-shot or few-shot setting, which
eliminates the need for large task-specific datasets. 
While in the zero-shot setting the pre-trained LM is used in combination with a natural language description of the task only, in the few-shot setting, the model additionally receives one or more examples of the task. Both settings rely on the \textit{in-context learning} ability of large LMs, i.e., their ability to perform a downstream task based on a prompt consisting of a natural language instruction and, optionally, a few additional task demonstrations~\cite{brown2020language}. 
While in-context learning describes the conditioning of a pre-trained model on a downstream task using a prompt, prompt engineering is concerned with finding effective prompts for in-context learning. 
Typically, prompt engineering involves the development of task-specific prompt templates, which describe how a prompt should be formulated to enable the pre-trained model to perform the downstream task at hand~\cite{liu2023pre}. 
Coming back to the activity-identification task in the context of transforming a text to a process model, such a prompt template could include a conceptual definition of ``Activity'', the task instruction ``Identify the activity:'', and examples of the task.
The development of such templates can be challenging, since the in-context learning performance of an LM is highly sensitive to the prompt format, including examples and the order of these examples \cite{perez2021true, zhao2021calibrate}.

\

%% file: sections/potentials.tex
\begin{figure}[t]
    \centering
    \includegraphics[width=0.9\linewidth]{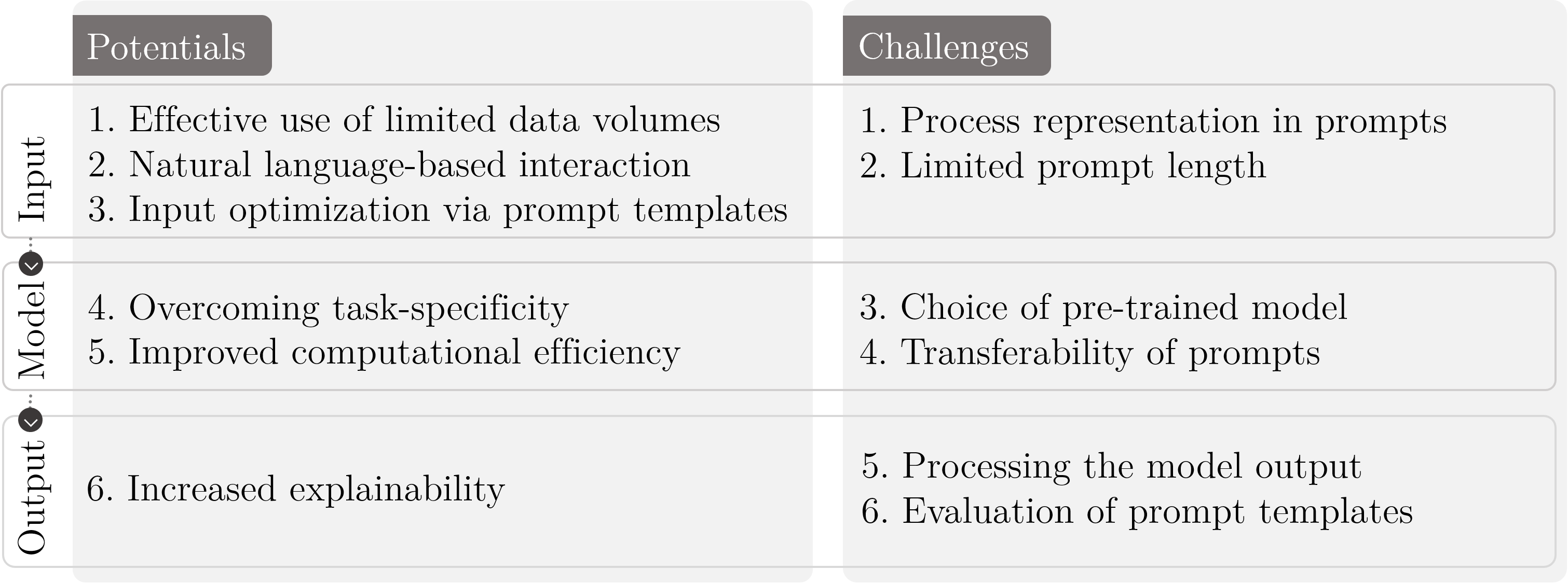}
    \caption{Potentials and challenges of prompt engineering for BPM}
    \label{fig:potentials}
\vspace{-1em}
\end{figure}

In this section, we discuss the potentials arising from the use of prompt engineering in BPM. Figure~\ref{fig:potentials} provides an overview of the six potentials we identified and whether they relate to the input (i.e., the use of prompts), the LM itself, or the output. In the following paragraphs, we discuss each potential in detail. 

\mypar{1. Effective use of limited data volumes} 
For many BPM tasks, the acquisition of large labeled datasets is difficult and costly, which can be attributed to several factors. First, annotating data for BPM tasks requires expert or at least domain knowledge~\cite{van2018challenges}. Second, there exists a vast variety of BPM tasks, each requiring separate labeled training datasets. Third, legal aspects limit the availability of data for both organizations and academia, as process-related data can contain sensitive information about the organizations' internal operations. As a result, researchers rarely have access to large amounts of high-quality process data from practice, forcing them to use LMs on small datasets.
Applying standard fine-tuning using such small datasets can result in poor performance, as many problems are difficult to grasp from just looking at a few examples~\cite{schick2020exploiting}. Prompt engineering circumvents these issues by integrating task specifications into the input sequence provided to the LM and has been shown to achieve competitive performance when compared to fine-tuning in low-data regimes~\cite{shin2020autoprompt, brown2020language, schick2020s}. This indicates a great potential in the BPM field, where limited availability of large task-specific datasets and poor data quality are common issues~\cite{kappel2021evaluating,bellan2022extracting}.

\mypar{2. Natural language-based interaction} 
In recent years, the NLP field has been characterized by a rapid pace of innovation. 
However, leveraging the breakthrough developments for BPM tasks has usually required specialized knowledge in deep learning. 
By leveraging natural language task specifications to employ LMs, prompt engineering has the potential to make these ever-more complex models readily accessible and customizable for BPM researchers and practitioners, regardless of their background. In particular, prompts represent an interpretable interface to communicate with LMs~\cite{brown2020language}, which enables incorporating expert knowledge into LMs by simply changing the instruction or examples contained in the prompt. For instance, prompts can provide information about the BPM domain, or, in the context of extracting process information from text, definitions of ``Activity'', ``Participant'', and other elements to be extracted~\cite{bellan2022extracting}.
While manually crafting appropriate prompts may still require domain expertise and experience with LMs, research on automated prompt generation~\cite{shin2020autoprompt} can be expected to further lower the barrier to leveraging LMs for BPM.

\mypar{3. Input optimization via prompt templates}
As already discussed, researchers in the BPM field often lack access to high-quality training datasets from practice. This can be problematic when fine-tuning an LM for a downstream task, as training on a dataset containing erroneous samples can lead to undesirable results. To illustrate this, consider the task of next-activity prediction in the context of predictive process monitoring. Now suppose that the training dataset contains process instances with semantic anomalies, e.g., a process instance in which an order is both accepted and rejected. When fine-tuning an LM on such a dataset, it may learn that this is correct process behavior and apply this in prediction tasks.
However, in a prompt, we could extend the provided instruction for the prediction task and tell the LM to ignore semantic anomalies in the provided examples. Thus, prompt engineering can help to mitigate issues with erroneous inputs by designing prompts, which enable the LM to use its general-purpose knowledge and correct erroneous examples itself.

\mypar{4. Overcoming task-specificity}
The fine-tuning approach achieves strong performance by transforming a pre-trained LM into a task-specific model using an individual set of annotated samples for every new task. However, the task-specificity of the models is a major limitation, since this means that a new model needs to be trained for each BPM task. As an example, consider the two tasks of transforming a process model into text and predicting the next activity of a running process instance. In the past, these tasks would have been addressed by two completely different, potentially highly specific techniques.
Prompt engineering can address this limitation.          
Given appropriate prompts, a single LM can be used across a wide range of tasks~\cite{brown2020language}. 
Thus, prompt engineering can help to use task-agnostic models and develop methods that can be applied across different BPM tasks.
 
\mypar{5. Improved computational efficiency} 
Increasing the size of LMs has been shown to result in performance improvements across a wide range of NLP tasks, particularly in settings that use in-context learning~\cite{brown2020language}. 
In recent years, LMs have grown from less than 350 million parameters~\cite{devlin-etal-2019-bert} to more than 175 billion parameters~\cite{brown2020language}. 
Fine-tuning such large models requires substantial amounts of time and computational resources, which limits the models’ accessibility and results in an immense carbon footprint. 
Prompt engineering, in contrast, is a fast and more sustainable approach of using a pre-trained model for a downstream task, reducing the cost of tailoring large models to new applications. Prompt engineering can thus help organizations to leverage the general-purpose abilities of a pre-trained LM for the management of their operations in a more timely, cost-effective and sustainable manner.

\mypar{6. Increased explainability} 
BPM supports decision-making in organizations, also in critical domains such as healthcare or financial services, which makes it essential to understand the rationale of employed systems. Therefore, the explainability and interpretability of artificial intelligence is becoming a growing area of interest in the BPM field~\cite{galanti2020explainable}. 
Prompt engineering can contribute to this emergent research direction in several ways. 
First, prompts provide an interpretable window into the task-specific behavior of an LM, since they contain all information about a downstream task that the model obtains. In contrast, the quality of the data used for fine-tuning may be less transparent to an LM's user. 
Second, prompt engineering can help LMs to decompose a task into intermediate steps, which can help to understand how the LM arrived at a particular output. 
In addition, the decomposition of a task allows for debugging in case of incorrect outputs and can improve the overall performance of an LM~\cite{wei2022chain}.  
Consequently, prompt engineering has the potential to foster more trust in LMs and increased LM adoption by BPM researchers and practitioners.

%% file: sections/challenges.tex
To realize the outlined potentials, various challenges have to be overcome.  
Figure~\ref{fig:potentials} provides an overview of the challenges we identified and whether they relate to the input (i.e., the use of prompts), the LM itself, or the output. 

\mypar{1. Process representation in prompts}  
While prompts are task specifications provided in natural language, the input for many BPM tasks is rarely simple text but includes potentially complex representations such as process models or event logs. In addition, many BPM tasks, e.g., process model matching~\cite{mendling201525}, face the challenge of dealing with process data of varying levels of abstraction. 
This raises the question of how process models or event logs can be expressed in a prompt in such a way that they can be effectively processed by an LM.
As an example, consider the task of transforming process models into textual descriptions~\cite{van2018challenges}. We could leverage an LM for this task by representing the process model in a prompt and asking the LM to give a description of the process. Finding such representations for complex process models, which contain sophisticated process structures like gateways or pools, is challenging, as it is not obvious how the non-sequential complexity of process models can be captured in a prompt. The development of prompt templates that are able to incorporate complex process representations and cover a large span of abstraction levels
is thus a significant challenge for leveraging LMs for BPM tasks trough prompt engineering.

\mypar{2. Limited prompt length}
The limited input length of an LM restricts the amount of context and instructions that can be provided for a downstream task in a single prompt, making it difficult to include extensive information or numerous task demonstrations. This poses a particular challenge in the BPM field, as process representations often contain a lot of different information, such as resources, responsibilities, or types of activities. Simply using the XML-format of a process model in a prompt, for example, will most likely exceed the input length limitation. When developing prompt templates, it is thus essential to carefully select the pieces of information contained in a process representation that are important for the specific BPM task and need to be provided to the LM via prompt. As an example, consider the process model autocompletion task~\cite{mendling201525}. When recommending the next modeling step at a user-defined position in a process model, elements close to the given position may be more relevant for this task than other elements appearing in the process model \cite{sola2021rule}. Thus, when developing prompt templates for BPM tasks under the restriction of limited prompt lengths, such additional considerations need to be taken into account.

\mypar{3. Choice of pre-trained model}  
In recent years, a number of pre-trained LMs have been published. Notable examples include BERT~\cite{devlin-etal-2019-bert}, GPT-3~\cite{brown2020language}, and T5~\cite{raffel2020exploring}. While several pre-trained LMs have already been employed in the BPM field, there is a lack of systematic comparisons of the existing process knowledge that is contained in these models. Similar to the evaluation of commonsense in pre-trained models \cite{zhou2020evaluating}, it could be beneficial for the selection of a pre-trained model to evaluate the process knowledge contained in different LMs. However, such an evaluation requires benchmarks that are currently not available. Similar benchmarks are also needed for a systematic comparison of the benefits that prompt engineering provides for different pre-trained LMs and BPM tasks.

\mypar{4. Transferability of prompts}
The development of LMs is constantly advancing, with new models, trained on larger datasets or with new training techniques, appearing frequently.
Against this background, it is essential to understand the extent to which selected prompts are specific to the LM. A first study that investigates the transferability of prompts for different NLP tasks and unsupervised open-domain question answering has been conducted by Perez et al.~\cite{perez2021true}. Their study shows that prompt transferability is poor when the model sizes are different. In other words, the same prompt can lead to performances discrepancies between LMs of different sizes. 
Therefore, similar studies for BPM tasks are needed in order to learn about the transferability of prompts across different LMs in the BPM field and to make prompt engineering efficient in the long run.

\mypar{5. Processing the model output}
When a model is presented with a prompt, it generates an output in response. However, it may be necessary to conduct a post-processing step to convert the model output into a format that is conducive for the BPM task at hand. To illustrate this, consider the translation of process models from a source into a target language~\cite{van2018challenges}. 
Prompt engineering for this task involves the transformation of a given process model into a textual representation that can be processed by an LM to generate a translation. However, obtaining a translated process model from the translated textual representation requires an additional step.
This post-processing step, which addresses the model output, can be challenging as it requires both domain knowledge and an understanding of how the LM generates output.

\mypar{6. Evaluation of prompt templates}
Process representations, such as process models or event logs, can be incorporated into a prompt in many different ways. For example, in the case of an event log, the events could be separated by ``,'' or ``then''. For more complex process representations, a high number of possible prompt templates can be expected. Taking into account that LMs are highly sensitive to the choice of prompts~\cite{zhao2021calibrate}, even similar variants of prompt templates should be evaluated. 
Therefore, instead of a brute-force trial and error approach with the prompt, it is essential to develop systematic ways for the evaluation of different prompt templates. 
A complementary approach to systematic evaluations could be the reduction of evaluation effort by disregarding prompt templates that do not adhere to design guidelines for prompt engineering in the BPM field. This would necessitate research on such guidelines, similar to those for prompt engineering for image generation~\cite{liu2022design}.
The evaluation of prompt templates thus represents a challenge for prompt engineering in the BPM field, which can be addressed through different research directions.

%% file: sections/conclusion.tex
In this paper, we examined the potentials and challenges of prompt engineering for BPM, providing a research agenda for future work. 
We demonstrated that the shift \textit{from fine-tuning to prompt engineering} can enhance research efforts in the application of LMs in BPM.
While our work should not be seen as prescriptive nor comprehensive, we expect it to help in positioning current research activities and in fostering innovative ideas to address the identified challenges.